\documentclass[conference]{IEEEtran}
\IEEEoverridecommandlockouts
\usepackage{cite}
\usepackage{amsmath,amssymb,amsfonts}
\usepackage{algorithmic}
\usepackage{graphicx}
\usepackage{textcomp}
\usepackage{xcolor}
\usepackage{booktabs}
\usepackage{multirow}
\usepackage{url}
\def\BibTeX{{\rm B\kern-.05em{\sc i\kern-.025em b}\kern-.08em
    T\kern-.1667em\lower.7ex\hbox{E}\kern-.125emX}}
\begin{document}

\title{Closing the Quality Gap in Low-Resource Text-to-Speech: LoRA Fine-Tuning of VoxCPM2 for Khmer and Korean}

\author{\IEEEauthorblockN{Phannet Pov}
\IEEEauthorblockA{\textit{Dept.\ of Big Data} \\
\textit{Chungbuk National University}\\
Cheongju-si, South Korea \\
phannetpov@chungbuk.ac.kr
}
\and
\IEEEauthorblockN{Hyun Woo Park}
\IEEEauthorblockA{\textit{Dept.\ of Big Data} \\
\textit{Chungbuk National University}\\
Cheongju-si, South Korea \\
hyunwoo50@chungbuk.ac.kr}
\and
\IEEEauthorblockN{Voneat Pen}
\IEEEauthorblockA{\textit{Dept.\ of Digital Transformation and Research} \\
\textit{CamTech University}\\
Phnom Penh, Cambodia \\
voneat.pen@camtech.edu.kh}
\and
\IEEEauthorblockN{Sovandara Chhoun}
\IEEEauthorblockA{\textit{Dept.\ of Big Data} \\
\textit{Chungbuk National University}\\
Cheongju-si, South Korea \\
sovandara@chungbuk.ac.kr}
\and
\IEEEauthorblockN{Wan-Sup Cho}
\IEEEauthorblockA{\textit{BigDataLabs Co., Ltd.} \\
\textit{Dept.\ of Management Information Systems} \\
\textit{Chungbuk National University}\\
Cheongju-si, South Korea \\
wscho@chungbuk.ac.kr}
\and
\IEEEauthorblockN{Saksonita Khoeurn*}
\IEEEauthorblockA{\textit{BigDataLabs Co., Ltd.} \\
\textit{Dept.\ of Management Information Systems} \\
\textit{Chungbuk National University}\\
Cheongju-si, South Korea \\
saksonita@chungbuk.ac.kr}
}

\maketitle

\begin{abstract}
Large pretrained text-to-speech (TTS) models sound almost human for well-resourced languages, but much worse for languages that are rare in their training data. We study this quality gap for Khmer and Korean using VoxCPM2, a 2.4B-parameter, tokenizer-free TTS model that joins a MiniCPM-4 language-model backbone with a flow-matching diffusion decoder. We build one shared, language-tagged corpus of 25.5 hours after cleaning and adapt VoxCPM2 with a single Low-Rank Adaptation (LoRA) adapter, trained on both languages at once and added to both the language model and the decoder. The adapter is zero-initialized, so training starts exactly at the original zero-shot model. In native-speaker listening tests, the Khmer Mean Opinion Score (MOS) rises from 3.85 to 4.23 with the best adapter, rank 64. This gain is highly significant under a paired Wilcoxon test with {\boldmath$p<0.001$}, and it is achieved while training only 0.19 to 3.03 percent of the parameters. Two findings stand out. First, the training loss and human ratings disagree on the best rank. The loss is lowest at rank 128, but MOS peaks at rank 64. Second, the same adapter gives no significant gain for Korean, which the base model already covers well, and a high rank even hurts quality. This shows that adaptation helps mainly where the base model is truly weak.
\end{abstract}


\section{Introduction}
Neural text-to-speech (TTS) has improved quickly. Early systems predicted acoustic features, such as Tacotron~2~\cite{tacotron2} and FastSpeech~2~\cite{fastspeech2}. Newer end-to-end and large generative systems~\cite{vits,valle} now sound close to human for well-resourced languages such as English and Mandarin. The main reason is large-scale pretraining. Foundation models such as VoxCPM~\cite{voxcpm} and its successor VoxCPM2~\cite{voxcpm2} are trained on millions of hours of multilingual speech and synthesize highly natural, context-aware speech.

Even so, this speech is not yet equal to a real human speaker. The base model is already useful, and its zero-shot output is reasonably good, but it still mispronounces some words, has unnatural prosody, and sounds less natural than a native speaker. This shortfall is larger for under-resourced languages that appear rarely in pretraining~\cite{mms,fleurs}, and our goal is to close this remaining gap between the model and natural human speech. We study Khmer, the official language of Cambodia and a genuinely low-resource language; unlike English or Korean, Khmer is written without spaces between words~\cite{khmerseg}. To test whether adaptation helps every language or only those with a real deficit, we pair Khmer with Korean, a language that the base model already covers well.

The usual fix is full fine-tuning, which updates every parameter. It can restore quality, but it is costly because it needs a lot of compute and storage, a separate copy of a multi-billion-parameter model for each language, and it can make the model forget what it already knew. Parameter-efficient fine-tuning (PEFT) avoids this. In particular, Low-Rank Adaptation (LoRA)~\cite{lora} keeps the pretrained weights frozen and trains only small low-rank matrices, so only a tiny fraction of the parameters change. LoRA is well proven for large language models~\cite{lora,qlora}. For speech, two questions are still open. The first is how well it closes the quality gap for low-resource TTS, and the second is whether a single shared adapter can serve several very different languages at the same time.

This paper investigates whether a small LoRA adapter on VoxCPM2 can close this gap, using Khmer and Korean, two languages that the base model covers to different degrees. Our contributions are:

\begin{itemize}
\item \textbf{One shared adapter for two languages and two modules.} We train a single LoRA adapter on Khmer and Korean together, and we add it to both the MiniCPM-4 language model and the flow-matching decoder. One small adapter, which trains only $0.19$ to $3.03\%$ of the parameters, then serves both scripts, with no separate model per language. As far as we know, this is the first parameter-efficient adaptation of a foundation TTS model for Khmer.
\item \textbf{Adaptation helps only where the base model is weak.} We measure native-speaker MOS and test for significance. The same adapter gives a large, highly significant gain for Khmer, which the base model covers poorly, as overall MOS rises from 3.85 to 4.23, an improvement of 0.38 points with $p<0.001$. It gives no significant gain for Korean, which the base model already covers well, since the best rank is only 0.11 points higher, with $p=0.49$. A high rank even makes Korean worse. So the adapter fills a real gap, rather than helping every language equally.
\item \textbf{Training loss does not predict the best rank.} We test ranks 8, 16, 32, 64, and 128. The validation loss is lowest at rank 128, but Khmer MOS is highest at rank 64 for naturalness, prosody, and pronunciation, and then drops. So the loss overstates the value of extra capacity. The rank should be chosen by listening tests, and the small rank-8 adapter already recovers most of the gain.
\item \textbf{Adaptation as a simple probe of what the model already knows.} Because the adapter starts at the exact zero-shot model, how much it helps, and whether it helps at all, shows how much of a language the base model already learned. The useful rank grows with this gap. This gives clear advice to use rank 64 for Khmer, to avoid fine-tuning Korean for overall quality, and to avoid rank 64 or higher for Korean. It also shows that one global rank is wrong when languages differ.
\end{itemize}

\section{Related Work}

\subsection{Neural Text-to-Speech}
Modern TTS began with two-stage neural pipelines. Tacotron~2~\cite{tacotron2} predicts mel-spectrograms autoregressively and pairs them with a neural vocoder, while FastSpeech~2~\cite{fastspeech2} introduced non-autoregressive synthesis with explicit duration, pitch, and energy modeling. Fully end-to-end systems such as VITS~\cite{vits} combine variational inference with adversarial training and normalizing flows. Large generative models then recast TTS as a conditional language-modeling or diffusion problem, and VALL-E~\cite{valle} frames zero-shot TTS as neural-codec language modeling. VoxCPM~\cite{voxcpm} departs from discrete-codec approaches with a tokenizer-free design that models continuous acoustic representations, and VoxCPM2~\cite{voxcpm2} scales this to a 2.4B-parameter model that pairs a MiniCPM-4~\cite{minicpm} backbone with a flow-matching~\cite{flowmatching} diffusion decoder. We adopt VoxCPM2 as our base model.

\subsection{Multilingual and Low-Resource TTS}
Extending speech technology to under-resourced languages is a long-standing challenge. Massively multilingual efforts such as the Massively Multilingual Speech project~\cite{mms} and the Common Voice corpus~\cite{commonvoice} expand language coverage. For zero-shot and cross-lingual TTS, YourTTS~\cite{yourtts} and XTTS~\cite{xtts} demonstrate transfer to new speakers and languages from limited data. Nevertheless, languages such as Khmer remain data-scarce, and quality for nominally supported low-resource languages typically lags that of high-resource languages, motivating targeted adaptation.

\subsection{TTS Adaptation}
A body of work adapts pretrained TTS models to new speakers, styles, or languages from limited data~\cite{voicecloning}. AdaSpeech~\cite{adaspeech}, for example, adapts a model while updating only a small set of parameters, foreshadowing parameter-efficient approaches. These methods establish that high-quality adaptation need not retrain the full model. Our work extends this line to lightweight, jointly multilingual adaptation of a 2.4B-parameter foundation TTS model, where a single shared low-rank adapter serves two typologically distinct, under-resourced languages at once.

\subsection{Parameter-Efficient Fine-Tuning}
Parameter-efficient fine-tuning adapts large pretrained models while updating only a small subset of parameters. LoRA~\cite{lora} injects trainable low-rank matrices into frozen weight matrices, and QLoRA~\cite{qlora} further reduces memory via quantized backbones. PEFT is standard for adapting large language models and is increasingly applied to speech~\cite{peftspeech}. Its use for closing the quality gap in under-resourced TTS, and whether a single shared adapter can jointly serve multiple typologically distinct languages, has received limited attention; we address this directly.

\section{Methodology}

\begin{figure*}[!t]
\centering
\includegraphics[width=\textwidth,trim=15 207 86 5,clip]{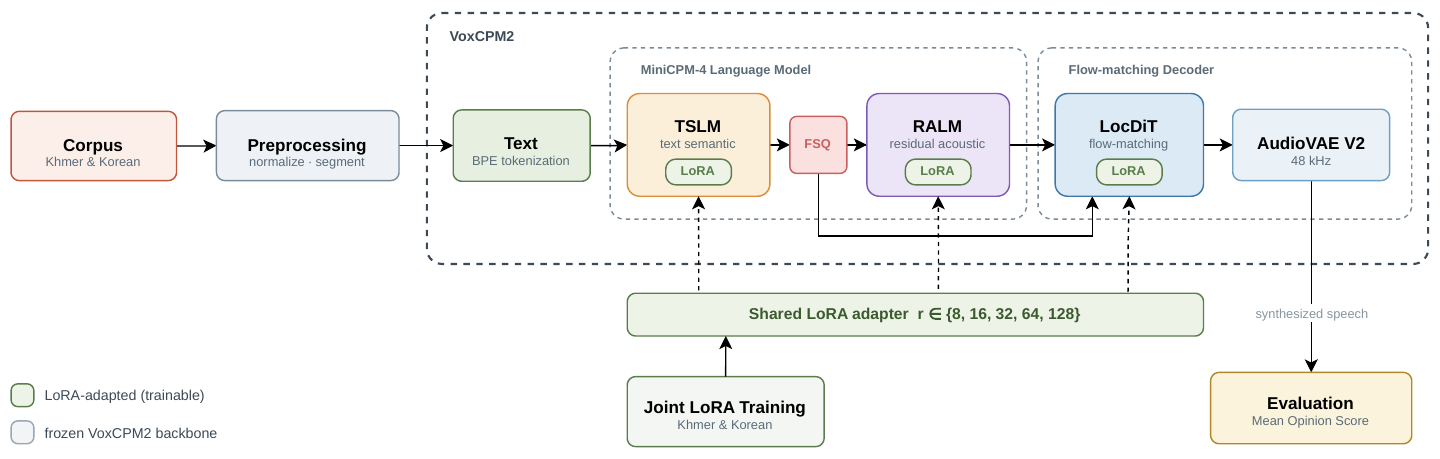}
\caption{Proposed shared-LoRA fine-tuning pipeline for VoxCPM2.}
\label{fig:pipeline}
\end{figure*}

\subsection{Model}
Figure~\ref{fig:pipeline} gives an overview of our pipeline. We build on VoxCPM2~\cite{voxcpm2}, a tokenizer-free TTS model with about $2.39\times10^{9}$ parameters. The input is a text prompt, which is normalized, segmented, and BPE-tokenized before entering the model. VoxCPM2 has two parts. The first is a MiniCPM-4~\cite{minicpm} language-model backbone with hidden size 2048, 28 transformer layers plus 8 residual layers, 16 attention heads with 2 key/value heads, and a vocabulary of 73{,}440. In this backbone, a text-semantic stage (TSLM), a finite scalar quantization stage (FSQ), and a residual-acoustic stage (RALM) turn the tokens into an acoustic representation. The second is a flow-matching~\cite{flowmatching} diffusion transformer (DiT) decoder, shown as a local DiT (LocDiT) followed by the AudioVAE~V2 vocoder. It produces continuous acoustic features with feature dimension 64 and patch size 4, and renders 48~kHz audio. Unlike systems that use discrete codec tokens~\cite{valle}, VoxCPM2 predicts continuous features directly, which avoids codec quantization artifacts.

\subsection{Corpus}
We build one corpus for Khmer (km) and Korean (ko) from public and in-house sources, summarized in Table~\ref{tab:data}. These are a Khmer corpus provided by the Institute of Digital Research and Innovation (IDRI), Cambodia; the Korean Single Speaker (KSS) corpus~\cite{kss}; and Korean Common Voice/FLEURS~\cite{commonvoice,fleurs}. We prepare the data in four steps. In Step 1, aggregation, we pair each clip with its transcript and measure its duration. In Step 2, cleaning, we drop clips shorter than $0.5$~s or longer than $20$~s and check that audio and text match. In Step 3, tokenization, we add a language tag, [km] or [ko], to the front of each transcript, encode it with the VoxCPM2 tokenizer, whose vocabulary has 73{,}440 entries, and drop clips with more than 256 text tokens. Steps 2 and 3 together remove $7.1\%$ of the Khmer clips and $0.7\%$ of the Korean clips. In Step 4, manifest construction, we split each language 90/10 into train and validation, then upsample, that is, repeat, the Khmer training clips $2.81\times$ until Khmer is $40\%$ of the training mix, to make up for its scarcity. We keep the validation split at its natural ratio, so the loss is measured fairly. Table~\ref{tab:data} reports utterances and hours at every stage. Its rows follow the four steps above. Raw is the aggregated corpus, and Cleaned is what survives Steps 2 and 3. Train, unique and Validation are the two halves of the 90/10 split, and Train, upsampled is the manifest we actually train on, which counts each repeated Khmer clip once per repetition. Because upsampling repeats existing audio rather than adding new material, that manifest spans $28.31$~h of rows drawn from only $22.95$~h of unique audio; we report both, and all corpus sizes quoted elsewhere in this paper refer to unique audio after cleaning.

\begin{table}[htbp]
\caption{Corpus Statistics at Each Preparation Stage.}
\centering
\small
\renewcommand{\arraystretch}{1.15}
\setlength{\tabcolsep}{6pt}
\begin{tabular}{lrrrr}
\toprule
\multirow{2}{*}[-2pt]{\textbf{Stage}} & \multicolumn{2}{c}{\textbf{Khmer}} & \multicolumn{2}{c}{\textbf{Korean}} \\
\cmidrule{2-3} \cmidrule{4-5}
& \textbf{Utts.} & \textbf{Hours} & \textbf{Utts.} & \textbf{Hours} \\
\midrule
Raw                            & 4{,}000 & 3.96 & 15{,}769 & 22.21 \\
Cleaned                        & 3{,}717 & 3.30 & 15{,}658 & 22.20 \\
\midrule
Train, unique                  & 3{,}345 & 2.96 & 14{,}092 & 19.99 \\
Train, upsampled               & 9{,}395 & 8.31 & 14{,}092 & 19.99 \\
Validation                     &    372  & 0.33 &  1{,}566 &  2.21 \\
\bottomrule
\end{tabular}
\label{tab:data}
\end{table}

\subsection{Joint Multilingual LoRA Adaptation}
Rather than fine-tuning all parameters, we attach a single shared LoRA~\cite{lora} adapter to the frozen backbone. For a pretrained weight matrix $W_0 \in \mathbb{R}^{d \times k}$, LoRA constrains the update to a low-rank product:
\begin{equation}
W = W_0 + \Delta W = W_0 + \frac{\alpha}{r} B A,
\nonumber
\end{equation}
Here $A \in \mathbb{R}^{r \times k}$ uses the Kaiming-uniform initialization and $B \in \mathbb{R}^{d \times r}$ is set to zero. So $\Delta W = 0$ at the start, and training begins exactly at the original zero-shot model. We add the adapter to the query, key, value, and output projections of attention in both the language model, covering its base and residual layers, and the DiT decoder. The feed-forward linears and the audio VAE stay frozen. We set $\alpha = 2r$ and try ranks 8, 16, 32, 64, and 128. This is $4.5$ to $72.4$ million trainable parameters, or $0.19$ to $3.03$ percent of the base model.

Importantly, we train one adapter on Khmer and Korean together, using the language-tagged data. So a single set of low-rank matrices learns both scripts. The language tag tells the model which language it is reading, so the adapter can share capacity but still keep the scripts apart. This means we do not need a separate model or adapter for each language.

\subsection{Training Configuration}
We train every adapter with the AdamW optimizer with $\beta_1{=}0.9$, $\beta_2{=}0.999$, and weight decay $0.01$. The peak learning rate is $1\times10^{-4}$, with a 200-step linear warmup and then cosine decay to zero. The effective batch size is 16, from a micro-batch of 4 with gradient accumulation of 4. We clip gradients at $1.0$ and use bfloat16 mixed-precision training; the audio VAE stays in float32. Each run is 10{,}000 steps, validated every 500 steps. We use one NVIDIA H200 GPU, and each rank takes about 2.6 hours at around 1.07 steps per second.

\subsection{Evaluation Metrics}
Our main automatic metric is the validation flow-matching loss, the diffusion objective measured on the held-out validation split. A lower value means a better fit to the target speech. We also track the stop-token loss. Because the adapter starts at zero, the loss at the start of training equals that of the zero-shot base model, so the drop during training shows how much of the gap the adapter closes. We also synthesize speech. For each rank and for the base, we generate the same Khmer and Korean sentences at 48~kHz. Finally, we run MOS listening tests for both languages, reported in Tables~\ref{tab:mos_ko} and~\ref{tab:mos}. For each language, five male and female native speakers rate every system on a 5-point scale along three axes, namely naturalness, prosody, and pronunciation, over 20 sentences. Let $r_{s,a,i}$ be the score that rater $i$ gives system $s$ on axis $a\in\{\mathrm{nat},\mathrm{pros},\mathrm{pron}\}$, and let $\bar{m}_{s,a}=\frac{1}{N}\sum_i r_{s,a,i}$ be its mean. The overall MOS is the mean of the three axes:
\begin{equation}
\mathrm{MOS}_s=\frac{1}{3}\sum_{a}\bar{m}_{s,a}.
\nonumber
\end{equation} We compare each system with the zero-shot base using a paired Wilcoxon signed-rank test. We also give 95\% confidence intervals from a bootstrap with 10{,}000 resamples and report inter-rater agreement with Krippendorff's $\alpha$. The two languages used different rater panels, so we compare systems only within a language, never across languages.

\section{Results and Discussion}

\subsection{Rank Sweep and Perceptual Results}
Table~\ref{tab:loss} shows the trainable fraction, adapter size, and validation loss at 10{,}000 steps for the base and each rank. Tables~\ref{tab:mos_ko} and~\ref{tab:mos} show the native-speaker MOS, by axis, for Korean and Khmer.

\textbf{Automatic loss.} The validation flow-matching loss drops from about $0.83$ for the zero-shot base to between $0.71$ and $0.73$ for every rank. The loss is lowest at rank 128, with a value of $0.7094$. Rank 8 has the second-lowest loss of $0.7243$, yet its adapter is only $18$~MB, about $15$ times smaller than the $277$~MB rank-128 adapter.

\textbf{Khmer quality.} LoRA gives a large gain that grows with rank and then peaks. Overall MOS rises from $3.847$ for the zero-shot base to $4.227$ at rank 64, a gain of 0.38 points that is highly significant with $p<0.001$. Ranks 32 and 128 are also significant, both with $p=0.001$, while ranks 8 and 16 are not, with $p=0.19$ and $p=0.07$ respectively. The gain shows up on all three axes, and the biggest single jump is in prosody, which rises from $3.76$ to $4.36$, the strongest effect in the study. This suggests the base model's main Khmer weakness was rhythm and intonation. Quality rises up to rank 64 and then falls to $4.083$ at rank 128, so capacity beyond rank 64 hurts perceived quality even though the loss keeps dropping. In short, the loss and human ratings agree that adaptation helps but disagree on the best rank. Rank 64 is best for people, and even rank 8, with an MOS of $3.910$, already recovers most of the gain.

\textbf{Korean quality.} The result is very different. No adapter significantly improves overall Korean MOS. The best mean is $3.757$ at rank 32, compared with $3.650$ for the base, and this difference is not significant with $p=0.49$. In addition, the confidence intervals for ranks 8 to 32 all overlap the baseline. The only significant overall change is negative. Rank 64 scores $3.480$, which is significantly below the base with $p=0.02$, as naturalness falls from $3.67$ to $3.23$. The only local improvement is pronunciation, which reaches $4.03$ at rank 32 while naturalness and prosody stay flat. Korean is already well covered by the base model's pretraining, so there is little gap to close. At a high rank the extra capacity overfits the small training set and damages skills the model already had. The stop-token loss converges below $0.05$ for all ranks.

\begin{table}[!t]
\caption{Effect of LoRA Rank on Adapter Size and Validation Flow-Matching Loss.}
\begin{center}
\small
\renewcommand{\arraystretch}{1.2}
\setlength{\tabcolsep}{6pt}
\begin{tabular}{lccc}
\toprule
\textbf{System} & \textbf{\% base} & \textbf{Size} & \textbf{Val.\ loss}$\downarrow$ \\
\midrule
LoRA $r{=}8$   & $0.19$ & $18$\,MB  & $0.7243$ \\
LoRA $r{=}16$  & $0.38$ & $35$\,MB  & $0.7334$ \\
LoRA $r{=}32$  & $0.76$ & $70$\,MB  & $0.7344$ \\
LoRA $r{=}64$  & $1.51$ & $139$\,MB & $0.7266$ \\
LoRA $r{=}128$ & $3.03$ & $277$\,MB & $\mathbf{0.7094}$ \\
\bottomrule
\end{tabular}
\label{tab:loss}
\end{center}
\footnotesize{The base has $2.39\times10^{9}$ params, all frozen except the adapter, and $\alpha{=}2r$.}
\end{table}

\begin{table}[!t]
\caption{Native-Speaker Mean Opinion Scores for Korean.}
\begin{center}
\small
\renewcommand{\arraystretch}{1.2}
\setlength{\tabcolsep}{3pt}
\begin{tabular}{lccccr}
\toprule
\textbf{System} & \textbf{Nat.}$\uparrow$ & \textbf{Pros.}$\uparrow$ & \textbf{Pron.}$\uparrow$ & \textbf{Overall}$\uparrow$ & \textbf{$p$} \\
\midrule
Zero-shot base & $3.67$ & $3.64$ & $3.64$ & $3.65{\pm}0.76$ & --- \\
\cmidrule{1-6}
LoRA $r{=}8$   & $3.59$ & $3.64$ & $3.95$ & $3.73{\pm}0.72$ & $.60$ \\
LoRA $r{=}16$  & $3.65$ & $3.53$ & $3.94$ & $3.71{\pm}0.79$ & $.80$ \\
LoRA $r{=}32$  & $3.61$ & $3.63$ & $\mathbf{4.03}$ & $\mathbf{3.76{\pm}0.75}$ & $.49$ \\
LoRA $r{=}64$  & $3.23$ & $3.49$ & $3.72$ & $3.48{\pm}0.81^{*}$ & $.02$ \\
LoRA $r{=}128$ & $3.27$ & $3.45$ & $3.91$ & $3.54{\pm}0.80$ & $.10$ \\
\bottomrule
\end{tabular}
\label{tab:mos_ko}
\end{center}
\footnotesize{Asterisks denote a statistically significant difference from the zero-shot base under a paired Wilcoxon signed-rank test, with $^{*}p<0.05$, where rank 64 is a significant \emph{decrease}. Unmarked rows are not significant. The highest mean is shown in boldface.}
\end{table}

\begin{table}[!t]
\caption{Native-Speaker Mean Opinion Scores for Khmer.}
\begin{center}
\small
\renewcommand{\arraystretch}{1.2}
\setlength{\tabcolsep}{3pt}
\begin{tabular}{lccccr}
\toprule
\textbf{System} & \textbf{Nat.}$\uparrow$ & \textbf{Pros.}$\uparrow$ & \textbf{Pron.}$\uparrow$ & \textbf{Overall}$\uparrow$ & \textbf{$p$} \\
\midrule
Zero-shot base & $4.00$ & $3.76$ & $3.78$ & $3.85{\pm}0.77$ & --- \\
\cmidrule{1-6}
LoRA $r{=}8$   & $3.97$ & $3.98$ & $3.78$ & $3.91{\pm}0.65$ & $.19$ \\
LoRA $r{=}16$  & $4.07$ & $3.93$ & $3.82$ & $3.94{\pm}0.66$ & $.07$ \\
LoRA $r{=}32$  & $4.12$ & $4.01$ & $3.97$ & $4.03{\pm}0.67^{**}$ & $.001$ \\
LoRA $r{=}64$  & $\mathbf{4.25}$ & $\mathbf{4.36}$ & $\mathbf{4.07}$ & $\mathbf{4.23{\pm}0.58}^{***}$ & ${<}.001$ \\
LoRA $r{=}128$ & $4.19$ & $4.15$ & $3.91$ & $4.08{\pm}0.63^{**}$ & $.001$ \\
\bottomrule
\end{tabular}
\label{tab:mos}
\end{center}
\footnotesize{Asterisks denote a statistically significant difference from the zero-shot base under a paired Wilcoxon signed-rank test, with $^{***}p<0.001$ and $^{**}p<0.01$. Unmarked rows are not significant. The best result is shown in boldface.}
\end{table}

\subsection{Qualitative Synthesis}
For each rank and for the base, we synthesize the same Khmer and Korean sentences at 48~kHz. By ear, the adapted models pronounce Khmer subscript consonant clusters and Korean sound boundaries more reliably than the base, and their rhythm is steadier. This matches the lower validation loss. These samples come with the released artifacts.

\subsection{Discussion}
We draw three lessons. First, one shared adapter, trained on both languages with language tags and Khmer upsampling, adapts VoxCPM2 to two very different scripts at once, with no per-language model. Because it starts at zero, it begins at the zero-shot base and lowers the validation loss for every rank. Second, the loss is not a perfect guide to perceived quality. The loss is lowest at rank 128, but Khmer MOS is highest at rank 64, reaching $4.227$, and drops at rank 128. So rank 64 is best for listeners, and the tiny rank-8 adapter already recovers much of the gain with an MOS of $3.910$, which is handy for deployment and for swapping adapters at run time. Third, and most important, the benefit depends on the language. The same adapter gives a large, significant gain for Khmer, an improvement of 0.38 points with $p<0.001$, but no significant gain for Korean, where a high rank even hurts. The simplest explanation is that the useful adapter size grows with the gap between the language and what the base model already knows. When the gap is large, as for Khmer, the model can use more capacity, up to rank 64. When the gap is small, as for Korean, extra capacity has little to add and a high rank overfits. So adaptation should target languages where the base model is truly weak, and a single global rank does not fit all languages.

\subsection{Limitations}
Our tests have some limits. We now report MOS for both Khmer and Korean in Tables~\ref{tab:mos_ko} and~\ref{tab:mos}, but a full fine-tuning upper bound is left for future work. We also have not yet isolated the value of sharing. To show that one joint adapter beats separate per-language adapters, we would need a comparison at the same parameter budget, which we leave for future work. The ratings are also noisy. Inter-rater agreement is low, with Krippendorff's $\alpha$ of $0.31$ for Khmer and $0.26$ for Korean, so the Korean null result is especially uncertain. Each panel has only five raters and 20 sentences per system, and the two languages used different panels, so only within-language trends are valid, not absolute cross-language scores. The corpus is small. It has $25.5$~h after cleaning, of which only $22.95$~h of unique audio reach training, and the KSS Korean data comes from a single speaker. In addition, the Khmer gains come partly from upsampling, not from more unique data, since the Khmer training set spans just $2.96$~h of unique speech. Finally, the two languages differ not only in pretraining coverage but also in fine-tuning data source. Korean uses open public corpora, namely KSS and Common~Voice/FLEURS, while Khmer uses a private in-house corpus. The two sets may differ in recording quality, speaker variety, and amount. So the Khmer-vs-Korean difference mixes two things, namely how far each language is from what the base model knows, and the fine-tuning data itself. We do not fully separate these. This is another reason our claims hold within a language, not as a controlled comparison across languages.

\section{Conclusion}
We studied the quality gap that large pretrained TTS models show for low-resource languages, using Khmer and Korean with VoxCPM2. We built one language-tagged corpus from several sources and trained a single shared LoRA adapter on both languages, adding low-rank updates to both the language model and the decoder. This lowered the validation loss from about $0.83$ for the zero-shot base to as low as $0.709$. It also raised Khmer MOS from $3.85$ to $4.23$, an improvement of 0.38 points with $p<0.001$, while training only $0.19$ to $3.03$ percent of the parameters. The rank sweep shows that the loss is lowest at rank 128, but Khmer MOS is highest at rank 64, and even rank 8 recovers much of the gain. For Korean, which the base model already covers well, no adapter improves overall quality and a high rank makes it worse. This difference between languages shows that small LoRA adapters help most where the base model is truly weak. Future work will add a full fine-tuning comparison, use larger rater panels, and cover more languages.

\section*{Acknowledgment}
This work was supported by the Chungbuk National University BK21 program in 2026. The authors would like to thank the Institute of Digital Research and Innovation (IDRI), under the Cambodia Academy of Digital Technology (CADT), Cambodia, for providing the Khmer speech dataset used in this study.

\end{document}